\documentclass[journal, 10pt,twocolumn]{IEEEtran}
\usepackage{epsfig}
\usepackage{subfigure}
\usepackage{amssymb}
\usepackage{amsbsy}
\usepackage{amsmath}
\usepackage{cite}
\usepackage{url}
\usepackage{color}
\usepackage{float}
\usepackage{times,amsmath,epsfig}
\usepackage{xspace,latexsym,syntonly}
\usepackage{amssymb}
\usepackage{amsfonts}
\usepackage{textcomp}
\usepackage{subfigure}
\usepackage{amsbsy}
\usepackage{cite}
\usepackage[ruled]{algorithm2e}
\usepackage{algpseudocode}

\newtheorem{theorem}{Theorem}
\newtheorem{lemma}{Lemma}

\newtheorem{definition}{Definition}

\newcommand{\beq}{\begin{equation}}
\newcommand{\eeq}{\end{equation}}
\newcommand{\bea}{\begin{array}}
\newcommand{\ena}{\end{array}}
\newcommand{\bds}{\begin {itemize}}
\newcommand{\eds}{\end {itemize}}
\newcommand{\bdf}{\begin{definition}}
\newcommand{\blm}{\begin{lemma}}
\newcommand{\edf}{\end{definition}}
\newcommand{\elm}{\end{lemma}}
\newcommand{\bthm}{\begin{theorem}}
\newcommand{\ethm}{\end{theorem}}
\newcommand{\bprp}{\begin{prop}}
\newcommand{\eprp}{\end{prop}}
\newcommand{\bcl}{\begin{claim}}
\newcommand{\ecl}{\end{claim}}
\newcommand{\bcr}{\begin{coro}}
\newcommand{\ecr}{\end{coro}}
\newcommand{\bquest}{\begin{question}}
\newcommand{\equest}{\end{question}}

\newcommand{\larrow}{{\larrow}}

\def\urltilda{\kern -.15em\lower .7ex\hbox{\~{}}\kern .04em}

\begin{document}
\title{PoPS: Policy Pruning and Shrinking for \\Deep Reinforcement Learning}

\author{Dor Livne and Kobi Cohen
\thanks{$\copyright$ 2020 IEEE. Personal use of this material is permitted. Permission from IEEE must be obtained for all other uses, in any current or future media, including reprinting/republishing this material for advertising or promotional purposes, creating new collective works, for resale or redistribution to servers or lists, or reuse of any copyrighted component of this work in other works.}
\thanks{Dor Livne and Kobi Cohen are with the School of Electrical and Computer Engineering, Ben-Gurion University of the Negev, Beer Sheva 8410501 Israel. Email:dorliv@bgu.ac.il, yakovsec@bgu.ac.il}
\thanks{This work was supported in part by the U.S.-Israel Binational Science Foundation (BSF) under grant 2017723, and by the Cyber Security Research Center at Ben-Gurion University of the Negev under grant 076/16.}
}

\date{}
\maketitle

\begin{abstract}
\label{sec:abstract}
The recent success of deep neural networks (DNNs) for function approximation in reinforcement learning has triggered the development of Deep Reinforcement Learning (DRL) algorithms in various fields, such as robotics, computer games, natural language processing, computer vision, sensing systems, and wireless networking. Unfortunately, DNNs suffer from high computational cost and memory consumption, which limits the use of DRL algorithms in systems with limited hardware resources. 

In recent years, pruning algorithms have demonstrated considerable success in reducing the redundancy of DNNs in classification tasks. However, existing algorithms suffer from a significant performance reduction in the DRL domain. In this paper, we develop the first effective solution to the performance reduction problem of pruning in the DRL domain, and establish a working algorithm, named Policy Pruning and Shrinking (PoPS), to train DRL models with strong performance while achieving a compact representation of the DNN. The framework is based on a novel iterative policy pruning and shrinking method that leverages the power of transfer learning when training the DRL model. We present an extensive experimental study that demonstrates the strong performance of PoPS using the popular Cartpole, Lunar Lander, Pong, and Pacman environments. Finally, we develop an open source software for the benefit of researchers and developers in related fields.
\end{abstract}

\def\keywords{\vspace{.5em}
{\bfseries\textit{keywords}---\,\relax%
}}
\def\endkeywords{\par}
\keywords
Deep reinforcement learning, deep neural network, pruning algorithms.

\section{Introduction}
\label{ssec:intro}
Deep reinforcement learning (DRL) algorithms have attracted much attention in recent years due to their capability to provide a good approximation of the objective value in decision making tasks while dealing with very large state and action spaces. In contrast to classic reinforcement learning methods that perform well for small-size models but perform poorly for large-scale models, DRL combines a deep neural network (DNN) with reinforcement learning for overcoming this issue. The DNN is used to map from states to actions in large-scale models so as to yield a policy that maximizes the objective value. In DeepMind's recently published Nature paper \cite{mnih2013playing, mnih2015human}, a DRL algorithm was developed to teach computers how to play Atari games directly from the on-screen pixels, and strong performance was demonstrated in many tested games. In recent years, there is a growing attention of using DRL methods in various fields, such as robotics, natural language processing, computer vision, sensing systems, and wireless networking. A survey of recent studies can be found in \cite{li2017deep}.

The superior performance of DRL algorithms in decision making tasks has triggered the need to make them practically appealing when using cheap hardware devices. For example, industrialization of artificially intelligent engines in controlled manufacturing processes often requires small and cheap sensing devices to detect and respond to events. Another example is automatizing users in wireless communication and Internet of Things (IoT) systems which often consist of low power, computationally limited and battery constrained nodes. 

This issue has been recognized by the industry, and there are more and more players in the market that develop chips for low-power devices that support computationally intensive deep learning algorithms with low-power consumption. Prominent examples are Qualcomm Artificial Intelligence Engine, Intel's EyeQ family of system-on-chip (SoC) devices, Intel's Myriad 2 family, NXP's ADAS chip, and more. Along with these industrial developments, establishing fundamental algorithmic methods to reduce the size of DRL models is crucial for making them practically appealing for a wide range of applications that use systems with limited hardware resources. This challenge has triggered a new and fascinating research direction: \emph{How to train DRL models with strong performance while achieving compact representations of the DNNs?} In this paper we address this issue. 

\subsection{Contributions}
\label{ssec:contributions}

\begin{figure}[http]
\centering \epsfig{file=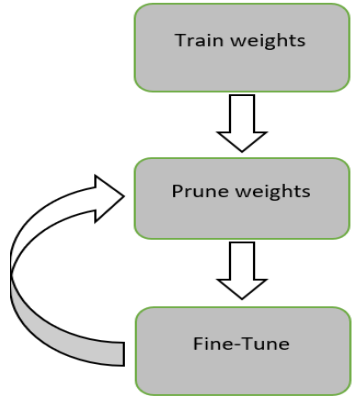, width=7cm, height=7cm}
\caption{An illustration of the pruning procedure. The first step is to train the model, the second is to prune it, and the last is to fine-tune the pruned model. The algorithm iterates the last two steps.}
\label{fig:prune_proc}
\end{figure}

In recent years, pruning algorithms have demonstrated considerable success in reducing the redundancy of DNNs in classification tasks (see related work in Section \ref{ssec:related}). The basic idea is to remove low-ranked neurons from the DNN which are redundant and do not contribute significantly to the output, resulting in a sparse DNN with a small number of non-zero parameters and faster operations. Pruning DNN is an iterative process, as illustrated in Fig. \ref{fig:prune_proc}. 
The popular vanilla version works as follows \cite{pruning}. The  first  step  is  to  train  the  model. Then, the algorithm enters an iterative process. At each iteration, the algorithm prunes the low-ranked connections with weights below a certain threshold. Then, it fine-tunes the model to recuperate from the pruning procedure. Eventually, the pruning procedure generates a sparse DNN, which is tractable for hardware implementations\cite{han2016eie}.

Despite the success of pruning methods in classification tasks, they suffer from a significant performance reduction in the DRL domain. There exist a number of recent studies that developed a DRL framework to guide the pruning process of a DNN \cite{ashok2017n2n, he2018amc}. However, those methods did not address the problem of training a DRL model with a compact representation as we consider here. The objective in \cite{ashok2017n2n, he2018amc} was to train DNNs with a compact representation used in classification tasks (and not in the DRL domain). The pruning actions were guided by a trained \emph{non-compact} DRL model. \emph{In  this paper, we develop the first effective solution to the performance reduction problem of pruning in the DRL domain, and establish an algorithm that achieves the state-of-the-art results in terms of training DRL models with strong required performance while minimizing the size of the DNN in the DRL domain.} Specifically, our contributions are summarized below:

1) \textbf{Developing a novel Policy Pruning and Shrinking (PoPS) algorithm for training DRL models:} In the DRL domain, the goal of an agent is to learn a policy, which is a mapping from a state to an action. Thus, the objective value at each state-action pair is not given as in supervised learning, but need to be learned online by exploring actions. Therefore, the agent faces the well known exploration versus exploitation dilemma. On the one hand, the agent should explore all actions in order to figure out their influence on the objective function. On the other hand, it should exploit the information gathered so far to choose the best actions. Existing pruning techniques do not operate well in the DRL domain since the ground truth is not given. Therefore, they must explore the state space in order to recuperate from the pruning procedure by interacting with the environment. This leads to a significant instability in the pruning procedure and increases the performance loss. As a result, when implementing pruning in the DRL domain, our algorithm must guard against losing a significant information regarding the optimal policy. Once the algorithm has this ability, it can detect redundancy in the DRL model, and train a regenerated shrunk dense model with strong performance. 

Building upon this insight, we develop the first effective solution to the performance reduction problem of pruning in the DRL domain. Specifically we establish a novel Policy Pruning and Shrinking (PoPS) algorithm to train DRL models with strong performance while achieving a compact representation of the DNN. PoPS executes an iterative procedure using three main steps to achieve this goal. In the first step, PoPS leverages the power of transfer learning to capture the full information regarding the desired policy. Specifically, PoPS trains a teacher network using a large-scale DRL model to yield a policy that maximizes the objective function without pruning. In the second step, PoPS executes a novel transfer learning-based \emph{policy pruning} procedure, which is controlled by the teacher, to find an efficient pruned representation of the model. The policy pruning step avoids the direct interaction with the environment when implementing pruning in the DRL domain. Since the teacher that controls the pruning process has already explored and exploited actions successfully, the fine tuning step has the ability to remove redundancy iteratively without the need of exploring actions which are less likely to contribute significantly to the objective function. In the third step, namely the \emph{policy shrinking} step, PoPS regenerates and trains a newly-constructed smaller dense model based on the redundancy measured by the policy pruning procedure. The policy pruning and policy shrinking steps are repeated until the algorithm can no longer detect any redundancy. A detailed description of PoPS algorithm is given in Section \ref{sec:PoPS}.

2) \textbf{Open source software}: We developed an open source software implementation of the PoPS algorithm. PoPS was developed using Python and is available at GitHub (see link in \cite{githubpops}). The open source software is built in an object oriented programming fashion which makes it easy to adjust to different environments and model settings. Our experimental study demonstrates the versatility of the software in four different challenging environments, as detailed in the next paragraph. We encourage the use of the open source software by researchers and developers in related fields.

3) \textbf{Achieving the state-of-the-art performance in extensive experimental study}: We evaluated our framework using four different challenging environments from the OpenAI gym library\cite{openai}, namely Cartpole, Lunar Lander, PONG, and Pacman environments, associated with different model architectures, namely deep Q-network (DQN) with a feed-forward fully connected network, Actor-Critic network with a feed-forward fully connected network, DQN with a Convolutional Neural Network (CNN), and dueling  DQN  with  CNN, respectively. We compared PoPS with the commonly used magnitude-based weight pruning \cite{pruning} using the gradual pruning framework \cite{pruneframework}, dubbed  Magnitude-Base Gradual Pruning (MBGP), which is known to achieve strong performance in many DNN architectures, and the Knowledge Distillation-Based Pruning (KDBP) algorithm which demonstrated improvements in the pruning procedure in various classification settings by using knowledge distillation when pruning DNNs \cite{yu2017accelerating}. \emph{In all tested environments, PoPS generated a compact representation of the DRL model with strong performance and a size of less than $1\%$ of the initial representation size. By contrast, the MBGP and KDBP algorithms were not able to reduce the size below $28\%$ in the Cartpole environment, $68\%$ in the Lunar Lander environemnt, $6\%$ in the Pong environment, and $8\%$ in the Pacman environment, of the initial representation size without significantly degrading the agent's performance. The results obtained by PoPS present the state-of-the-art performance in terms of training DRL models with strong required performance while minimizing the size of the DNN in the DRL domain.}
\subsection{Related Work}
\label{ssec:related}

DRL algorithms have attracted much attention in recent years due to their capability to provide a good approximation of the objective value while dealing with very large state and action spaces. In DeepMind's recently published Nature paper \cite{mnih2015human}, a DRL algorithm was developed to teach computers how to play Atari games directly from the on-screen pixels, and reported strong performance in many games. In \cite{foerster2016learning}, strong performance was found for several players in MNIST games and the switch riddle. A Double Q-learning method has been proposed in \cite{doubledqn}, where two DNNs simultaneously learn the policy and value evaluation. A survey of recent studies can be found in \cite{li2017deep}. This recent success has triggered the need to make DRL algorithms practically appealing when using cheap hardware devices. This is particularly relevant for DRL algorithms in sensing and inference systems (see \cite{wang2017active, puzanov2018derol_arXiv, puzanov2018derol_EAAI,  puzanov2018deep} and references therein), IoT, and wireless communication networks \cite{oshea2016deep, challita2017proactive, wangdeep, wang2018deep, naparstek2017deep_Globecom, naparstek2017deep, zhang2018deep}, that often require to operate using low power, computationally limited and battery constrained devices. The proposed PoPS framework in this paper addresses this challenge.

DNNs are typically over-parameterized, i.e, there is a significant redundancy in deep learning models \cite{denil}. This redundancy leads to a waste of both computation and memory usage, which limits their use in resource-constrained devices. Therefore, in recent years, various methods have been developed to find efficient compact representations of DNNs. In\cite{pruning}, the authors developed a pruning method followed by vector quantization and Huffman coding \cite{deepcompression} to compress neural networks. In \cite{trimming}, a pruning technique, called Network Trimming, was developed, where neurons are pruned based on the statistics of neuron activations. In \cite{channelpruning}, an iterative two-step algorithm was developed to effectively prune channels in each layer. In \cite{pruneframework}, the authors augmented Tensorflow with a pruning framework to prune the network connections during the training phase. In \cite{marculescu}, the authors developed a layer-compensated pruning algorithm for a layer-wise compensate filter pruning. In \cite{chin2019legr}, they proposed the LeGR algorithm, which is parameterized to learn layer-wise affine transformations over the filter norms. It constructs a learned global ranking to achieve an efficient resource-constrained filter pruning. 
In \cite{ashok2017n2n, he2018amc}, (non-compact) DRL models were used to explore the architecture space of the model in order to find an efficient architecture. Specifically, in \cite{ashok2017n2n}, DRL was used to determine which layers should be dropped and the size of the remaining layers. In \cite{he2018amc}, DRL was used to evaluate an efficient sparsity ratio of each layer. However, those methods did not address the problem of training a DRL model with a compact representation as we consider here. In \cite{quantization}, the authors developed the incremental network quantization (INQ), an iterative network quantization method that incorporates three interdependent operations: weight partitioning, group-wise quantization, and re-training. In \cite{ding2017lightnn, ding2018quantized}, LightNN algorithm that replaces the multiplications to one shift or a constrained number of shifts and adds in binarized NNs to improve the energy efficiency was developed and analzyed. Finally, the existing pruning methods suffer from a significant performance reduction in the DRL domain as explained and demonstrated in Sections \ref{ssec:contributions}, \ref{sec:experiemnts}.

Training a student network using the outputs of a trained teacher network was first introduced in \cite{bucilu2006model}, with the goal of compressing a large ensemble model to a single network. This approach was extended later in \cite{ba2014deep} to transfer knowledge from a deep neural network to a shallow one. More recently, it was used for policy distillation to train smaller dense networks that perform at the expert level \cite{policydist}. In this paper, we extend the method of training a student network to guard against the performance loss of pruning in the DRL domain. Transfer learning methods to fine-tune the model in the pruning procedure were used in \cite{lemaire2018structured,yu2017accelerating} in the form of Knowledge Distillation \cite{hinton2015distilling} to improve the stability and robustness of the pruning process in classification tasks. These methods, however, did not address the performance reduction problem of pruning in the DRL domain, which we address in this paper. 

\section{System Model and Problem Statement} 

\subsection{System Model}

\begin{figure}[htbp]
\centering \epsfig{file=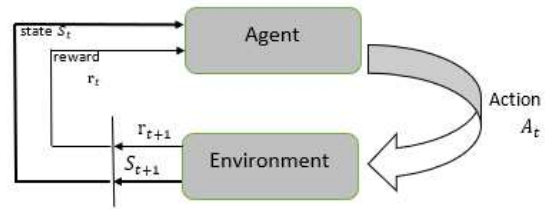,
width=0.58\textwidth}
\caption{An illustration of the system model. At each time (say $t$), the agent takes action $a_t\in \mathcal{A}$, receives reward $r_{t+1}$, and the system transits to state $s_{t+1}$, which is observed (or partially observed) by the agent.}
\label{RL}
\end{figure}

Consider a system consisting of an agent who interacts with an environment. Let $\mathcal{S}$ be the state space that the system can reach, and $s_t\in\mathcal{S}$ be the system state at time $t$. Let $\mathcal{A}$ be the action space that the agent can take. At each time (say $t$), the agent takes action $a_t\in \mathcal{A}$, receives reward $r_{t+1}$, and the system transits to state $s_{t+1}$, which is observed (or partially observed) by the agent. An illustration of the system is given in Figure \ref{RL}.

Let 
\begin{equation}
\displaystyle R = \sum_{t=1}^{T} \gamma^{t-1}r_t
\end{equation}
be the accumulated discounted reward, where \(0 \leq \gamma \leq 1\) is a discounted factor, and $T$ is the time horizon of the control problem. The agent's objective is to find a policy \(\pi : S \rightarrow A\) that maximizes the expected accumulated discounted reward:
\begin{equation}
\label{eq:max_R}
\displaystyle\max_{\boldsymbol{\pi}} \;\; \textbf{E}\left[R(\boldsymbol{\pi})\right],
\end{equation}
where $\textbf{E}\left[R(\boldsymbol{\pi})\right]$ denotes the expected accumulated discounted reward when the model performs policy $\boldsymbol{\pi}$.

\subsection{Background of Reinforcement Learning Solutions to (\ref{eq:max_R})}

Q-learning is a reinforcement learning method that aims at finding good policies to solve the dynamic programming problem in (\ref{eq:max_R}). It has been widely applied in various decision making problems, primarily because of its ability to evaluate the expected utility without requiring prior knowledge about the system model, and its ability to adapt when stochastic transitions occur \cite{watkins92}. The basic idea of Q-learning is to approximate the objective value (referred to as Q-value) for all possible state-action pairs. This is done by minimizing the Time Difference (TD) error using a stochastic approximation method that judiciously trades off between exploring all available actions to figure out their influence on the objective value, and exploiting actions which are likely to be more valuable in terms of maximizing the objective value. 

While Q-learning performs well when dealing with small action and state spaces, it becomes impractical when the problem size increases, mainly for two reasons: (i) A stored lookup table of $Q$-values for all possible state-action pairs is required, which makes the storage complexity intolerable for large-scale problems. (ii) As the state space increases, many states are rarely visited, which significantly decreases performance.

In recent years, DRL methods that combine DNN with Q-learning, have shown great potential for overcoming these issues. Using DRL, the DNN maps from the (partially) observed state to an action, instead of storing a lookup table of Q-values. Furthermore, large-scale models can be represented by the DNN well so that the algorithm can preserve good performance for very large-scale models. A well known DRL algorithm was presented in DeepMind's Nature paper \cite{mnih2015human} and demonstrated superior performance for teaching computers how to play Atari games directly from the on-screen pixels. Since then, there is a growing attention on using DRL methods in various fields (see related work in Section \ref{ssec:related}). 

\subsection{The Objective}

Unfortunately, DNNs are usually very demanding in computational power and memory usage, which limits the use of DRL algorithms in systems with limited hardware resources. Therefore, the objective of this paper is to develop a fundamental method to train DRL models with strong performance in terms of solving (\ref{eq:max_R}) while achieving compact representations of the DNNs. In the next section we address this challenge.

\section{The Policy Pruning and Shrinking (PoPS) Algorithm}
\label{sec:PoPS}

As discussed in Section \ref{ssec:contributions}.1, existing pruning algorithms do not effectively apply to the DRL domain, due to the performance loss with respect to the policy learning incurred by the pruning procedure when interacting directly with the environment. The suggested PoPS algorithm is designed to solve the issue of the performance reduction of pruning in the DRL domain. The framework is based on a novel iterative policy pruning and shrinking method that leverages the power of transfer learning when training the DRL model.

Note that running a sort of exhaustive search over layer sizes to find compact representations gradually becomes impractical in large-scale NNs due to the huge architecture space. The basic idea of PoPS is to use the power of pruning-based technique in detecting the redundancy over the DNN by dictating the compression direction in the huge architecture space. Thus, it avoids the need of running a laborious gradual search.

\subsection{Architecture of the PoPS Algorithm}
\label{ssec:popsarch}

\begin{figure*}[ht]
\centering\epsfig{file=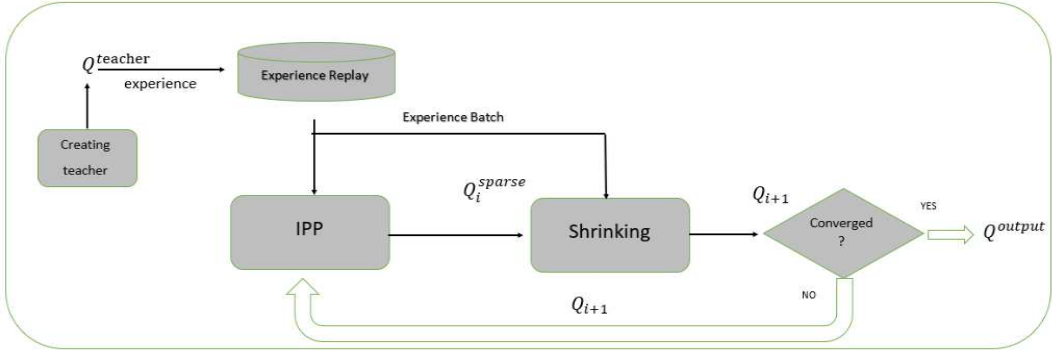,
width=16cm,height=6cm}
\caption{An illustration of the PoPS framework.}
\label{fig:pops}
\end{figure*}

In this section we present the architecture of PoPS algorithm. PoPS aims to find an efficient, compact, and redundant-free representation of the DNN. An illustration of PoPS architecture is given in Fig. \ref{fig:pops}. We next discuss in details each component in the architecture.\vspace{0.2cm} 
\begin{figure*}[ht]
\centering\epsfig{file=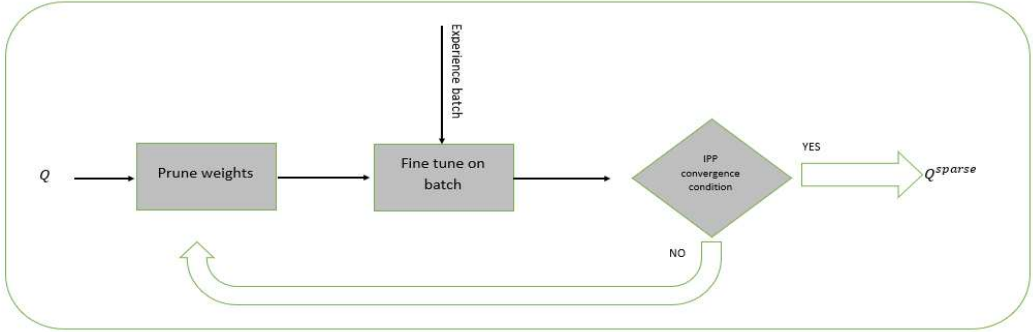,
width=16cm,height=6cm}
\caption{An illustration of the IPP module used in PoPS algorithm.}
\label{fig:ipp}
\end{figure*}
\begin{figure*}[ht]
\centering\epsfig{file=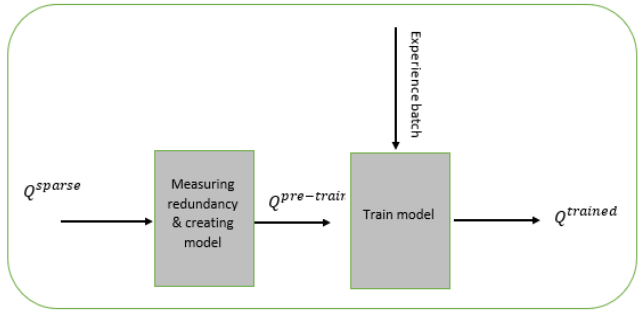,
width=11cm,height=6cm}
\caption{An illustration of the Shrinking module used in PoPS algorithm.}
\label{fig:shrink}
\end{figure*}

1) \emph{Creating a teacher:} This part of PoPS algorithm trains an agent to obtain a policy that solves (\ref{eq:max_R}) using a large-scale redundant DNN. The output of this module is the DRL model $Q^{teacher}$ with the corresponding DNN. This output is used as a teacher that guides the iterative policy pruning and shrinking steps. 

In order to train the teacher to solve (\ref{eq:max_R}), two commonly used training methods were incorporated in the framework. This makes the framework applicable to a wide range of decision making tasks. The first is a Q-learning type that uses the DNN (referred to as Deep Q-network or DQN) to map from the observed state to a Q-value for each action \cite{mnih2015human}. The action that maximizes the Q-value is selected (according to $\epsilon$-greedy distribution). Note that standard Q-learning methods use the same parameters for estimating the target and the current Q-value, resulting in a large correlation between the TD target and the updated parameters at every training step. This makes the training procedure highly unstable. Thus, the target DQN overcomes this issue by using a separate network with fixed parameters for estimating the TD target, where every $N$ steps the algorithm copies the parameters from the DQN network to the target network. 
Denote the DQN network by \(Q\) and the target network by \(Q^{target}\). Then, the updates to the Q-values satisfy the following equation:
\begin{equation}
\bea{l}
\label{eq:FixedQlearning}
\displaystyle
Q_{t+1}\left(s_t, a_t\right)=Q_t\left(s_t,a_t\right)
\vspace{0.2cm}\\\hspace{1cm}\displaystyle
+\alpha\left[r_{t+1}+\gamma\max_{a_{t+1}}Q^{target}_{t}\left(s_{t+1},a_{t+1}\right)\right.
\vspace{0.2cm}\\\displaystyle\left.\hspace{5cm}
-Q_t\left(s_t,a_t\right)\right], \vspace{0.0cm}
\ena
\end{equation}
where the subscript $t$ denotes the time index, and 
\beq
\label{td_value}
\displaystyle r_{t+1}+\gamma\max_{a_{t+1}}Q_t^{target}\left(s_{t+1},a_{t+1}\right) \vspace{0.0cm}
\eeq
is the learned value obtained by getting reward $r_{t+1}$ after taking action $a_t$ in state $s_t$, moving to the next state $s_{t+1}$, and then taking action $a_{t+1}$ that maximizes the future Q-value seen at the next state. 
The term $Q_t\left(s_t,a_t\right)$ is the old learned value. Thus, the algorithm aims at minimizing the TD error between the learned value and the current estimate value. The learning rate $\alpha$ is set to $0\leq\alpha\leq 1$, which is typically set close to zero.

The second commonly used method that we incorporate in the PoPS framework to train the teacher is the Advantage Actor-Critic method\cite{mnih2016asynchronous}. The actor critic model uses two neural networks. The first is an actor neural network that we denote by \(Q^\theta\), where \(\theta\) is the actor network parameters. The second is a critic network denoted by 
\(\tilde{Q}^w\), where \(w\) is the critic network parameters.
\(Q^\theta\) is used to determine the optimal action and \(\tilde{Q}^w\) is used to estimate the state-value function of the current state. Intuitively, the actor observes the environment reaction to it with an action, while the critic observes the actor and provides a feedback. 
\(Q^\theta\) and \(\tilde{Q}^w\) are trained simultaneously. The increments $\Delta w_t, \Delta \theta_t$ of the parameters $w, \theta$, respectively, at time $t$, satisfy the following equations:
\begin{equation}
\Delta w_t = \alpha (r_{t+1} + \gamma \tilde{Q}^w_t(s_{t+1}) - \tilde{Q}^w_t(s_{t}))\nabla_w \tilde{Q}^w_t(s_t), 
\end{equation}
and 
\begin{equation}
\Delta \theta_t = \nabla_{\theta} (\log(Q^{\theta}_t( s_t,a_t)))\tilde{Q}^w_t(s_{t+1}) .
\end{equation}

To further reduce the high variability in value-based methods, we use the advantage function instead of the value function such that:
\begin{equation}
\Delta \theta_t = \nabla_{\theta} (\log(Q^{\theta}_t( s_t,a_t)))A(s_t,a_t),    
\end{equation}
where
\begin{equation}
\label{eq:advantage_value}
A(s_t, a_t) = r_{t+1} + \gamma\tilde{Q}^w_t(s_{t+1})  - \tilde{Q}^w_t(s_{t+1}).
\end{equation}
This prodedure evaluates the gain in taking a specific action as compared to the average rewarded action at a given state. \vspace{0.2cm}

2) \emph{Experience Replay:} Once PoPS has trained the teacher, it periodically stores batches of randomly selected experience replay instances in the buffer (where typically the batch size is between $10,000$ and $30,000$). Note that for large-scale problems, the buffer capacity should not be too small to avoid missing important experiences and suffering from a performance reduction while replacing old experiences that reside in the buffer. In our experiments, the buffer capacity was set to $100,000$. Then, the iterative policy pruning and shrinking steps are trained by the experience replay instances and do not interact directly with the environment. It is important to note that the experience accumulation step is repeated until the student model has sufficient coverage of the state space for each training session. \vspace{0.2cm}

3) \emph{Iterative Policy Pruning (IPP):} We develop a fundamental iterative policy pruning (IPP) method to approach the performance limits when pruning the DNN, which are far to be reached by existing pruning methods in the DRL domain. IPP is an iterative process, as illustrated in Fig. \ref{fig:ipp}. At each iteration, IPP prunes the DNN by accumulating experiences from the teacher. It then fine tunes the model using policy distillation method\cite{policydist}. At each time step, IPP prunes each layer so that the sparsity of the layers satisfies the gradual pruning equation\cite{pruneframework}:
\begin{equation}
\begin{array}{l}
\displaystyle g_t = g^{\mbox{final}} + (g^{\mbox{initial}} - g^{\mbox{final}})\Big(1- \frac{t-t_0}{n\Delta}\Big)^3\;, \vspace{0.2cm}\\\hspace{3cm}
\displaystyle \text{ for \;\;} t\in\{t_0,\dotsc,t_0 + n\Delta t\},
\end{array}
\end{equation}
\normalsize
where \(t\) is the global step of the training process. The term \(g_t\) is the sparsity demand at the global step \(t\), \(g^{\mbox{initial}}\) is the initial sparsity of the layer (typically set to $0$), \(g^{\mbox{final}}\) is the target sparsity, and \(t_0\) is the global step from which the algorithm starts to prune (typically set to $0$). The term $n$ is the number of pruning steps to preform until \(g_t\) reaches \(g^{\mbox{final}}\), and \(\Delta\) is the pruning frequency.

Every $N$ time steps, IPP evaluates the model. If the evaluation of the model is lower then a predefined low threshold it stops the pruning, fine tunes the model using policy distillation with the KL-divergence loss function (see (\ref{kldiv}) below), and trains the model until it recuperates and solves (\ref{eq:max_R}). Specifically, Let \(D=\left\{s_i, q_{i}^t\right\}_{i=1}^N\)  be the state-value experience replay pairs generated by the teacher, and \(\left\{q_{i}^s\right\}_{i=1}^N\) be the student values (i.e., the model that we want to train). Then, IPP aims at matching the output distributions between \(\left\{q_{i}^s\right\}_{i=1}^N\) and \(\left\{q_{i}^t\right\}_{i=1}^N\) using the KL-divergence measure. Let \(\theta_s\) be the parameters of the student network. Then, \(\theta_s\) is trained to be the minimizer of the following loss function:
%
\begin{equation}
\label{kldiv}
\displaystyle L_{KL}(D,\theta_s) = \sum_{i=i}^{N} \mbox{softmax}\left(\frac{ q_i^t}{ \tau}\right)\log\left(\frac{\mbox{softmax}\displaystyle\left(\frac{q_i^t}{\tau}\right)}{\mbox{softmax}(q_i^s)}\right),
\end{equation}
where \(\tau\) denotes the temperature of the softmax function. 
If \(\tau\) is high, the output of the softmax function is softened. Since the outputs of the teacher are the expected future discounted reward of each possible action, IPP aims at making them sharper by taking a small \(\tau\).

Once the model recuperates, the pruning procedure continues. If the evaluation of the model solves (\ref{eq:max_R}), then IPP saves the model. The IPP continues accumulating experiences from the teacher and repeats this procedure until the convergence condition is met, i.e., the sparsity remains unchanged for $M$ consecutive iterations. Then, the output model is the one IPP saved which solves (\ref{eq:max_R}). 

\begin{figure}[htbp]
\centering \epsfig{file=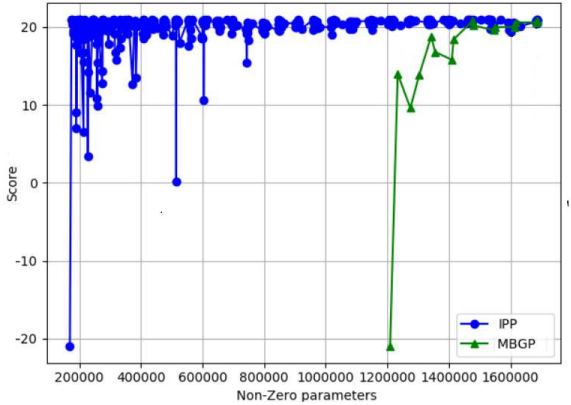,
width=0.5\textwidth}
\caption{The performance after one step of the iterative pruning procedure by IPP as compared to the iterative pruning procedure by the commonly used Magnitude-Based weight pruning \cite{pruning} using the Gradual Pruning (MBGP) framework \cite{pruneframework}.}
\label{fig:IPP_vs_vanilla}
\end{figure}

In Fig. \ref{fig:IPP_vs_vanilla} we illustrate the performance after one step of the iterative pruning procedure by IPP as compared to the iterative pruning procedure by the commonly used magnitude-based weight pruning \cite{pruning} using the gradual pruning framework \cite{pruneframework}. It can be seen that the IPP procedure achieves very good performance, and overcomes the significant performance loss typically incurred by the pruning procedure in the DRL domain.\vspace{0.2cm}

4) \emph{Policy Shrinking:} This part of PoPS algorithm shrinks the model to obtain a dense small-scale DNN based on the output from IPP. The first step is to measure the redundancy of the model based on the output from IPP, and generate the dense model accordingly. The second step is to train the newly generated model using experiences from the teacher. An illustration is given in Fig \ref{fig:shrink}. 

4.1) \emph{Measuring redundancy and creating model:} This step measures the number of non-zero parameters of the DNN output by the IPP module to determine the redundancy in that layer. This step determines the size of each layer which is required for efficient operations. Then, the algorithm shrinks the model, i.e., creates a dense DNN, where the number of variables of each layer is determined by the number of non-zero parameters that was measured in the sparse model. 

4.2) \emph{Train model:} 
This component trains the newly generated shrunken model by taking experiences from the teacher. The algorithm uses the policy distillation training method with KL-divergence loss function in this step as well, as detailed in the IPP module.

Finally, after the model is trained, the algorithm compares its size with the size of the model that was constructed at the previous iteration. If the difference is smaller then a predefined threshold it concludes that the algorithm has converged. Otherwise, it repeats the procedure. 

\subsection{Pseudo Code of the PoPS algorithm}

\begin{figure*}[ht]
\centering\epsfig{file=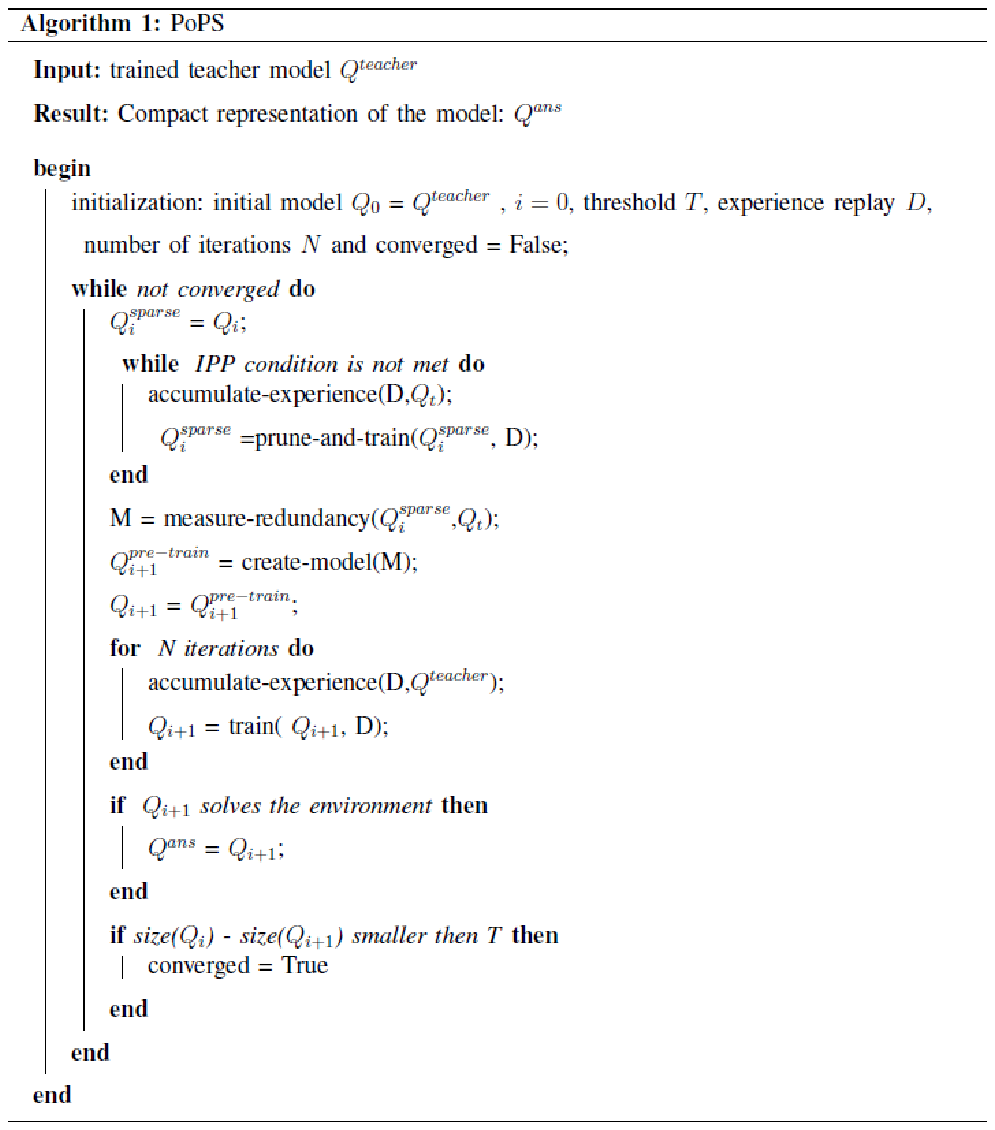,
width=14cm,height=15cm}
\label{pseudocode}
\end{figure*}

The pseudo code of PoPS algorithm is described in Algorithm 1. Let \(Q^{teacher}\) be a pre-trained model and let \(Q_i\) be a trained model at iteration $i$ (where \(Q_0\) = \(Q^{teacher}\) ). \(D=\left\{s_i, q_{i}^t\right\}_{i=1}^N\) is the Experience Replay set described in Section \ref{ssec:popsarch}. The description of Algorithm \ref{pseudocode} starts by pruning the model using the IPP module as detailed in Section \ref{ssec:popsarch}, and illustrated in Fig. \ref{fig:ipp}. IPP first accumulates experiences from \(Q^{teacher}\) stored in \(D\), as illustrated in Figure \ref{fig:pops} via the \textit{accumulate-experience} function in the pseudo code. Then, it prunes and fine-tunes the model using batches from $D$ and the KL-divergence loss function described in (\ref{kldiv}) by the \textit{prune-and-train} function. PoPS repeats this procedure until the IPP convergence condition is met, as detailed in Section \ref{ssec:popsarch}. Once the IPP convergence condition is met, the output of the IPP module \(Q_{i}^{sparse}\) is used as the input to the Shrinking module.

The Shrinking module measures the redundancy at each layer, denoted by $M$, by using the sparsity of \(Q_{i}^{sparse}\) via the \textit{measure-redundancy} function in the pseudo code. \(Q_{i+1}^{pre-train}\) denotes a randomly initialized model built by the measure $M$ via the \textit{create-model} function, which is then trained using batches from $D$ and the KL-divergence loss function described in (\ref{kldiv}) by the \textit{train} function. Eventually, a trained model, denoted by \(Q_{i+1}\), is stored as the output of the Shrinking module. PoPS ends when the size of the model converges.

\subsection{Open Source Software}
\label{ssec:open_source}
We developed an open source software implementation of PoPS algorithm. Researchers and developers in related fields are welcome to integrate PoPS in their working environment. PoPS was developed using Python and is available at GitHub (see link in \cite{githubpops}). The implementation details of the open source software are described in the Appendix.

\subsection{Complexity Discussion}
\label{ssec:complexity}
The number of multiplications through a feed-forward fully connected network with $G$ layers, in which $K$ is the size of the input state vector and \(d_g\) is the number of units in the \(g'th\) layer, is given by \(D = Kd_1 + \sum_{g=1}^{G-1}{d_g d_{g+1}}\). As a result, the computational complexity of the forward and back propagation for one sample is given by \(O(D)\). In PoPS, the complexity of accumulating \( N\) experiences by the teacher is \( O(DN)\) and the complexity of preforming back-propagation on \(M\) experiences is \(O(DM)\). Thus, each iteration of the IPP module has a complexity of order \(O(D(N+M))\). in the Shrinking module, given that at each iteration the model is trained on \(M \) experiences, the complexity of each iteration is \( O(D(N\ + M))\). Typically, the number of experiences is between $10,000$ and $30,000$. The number of iterations required to reach a compact architecture by PoPS depends on the amount of the redundancy in the initial DNN. Higher redundancy might result in running more iterations. In our experiments, PoPS converged to a compact architecture after no more then $9$ iterations, which is very efficient.

\section{Experiments}
\label{sec:experiemnts}

In this section we present an extensive experimental study to evaluate the performance of PoPS algorithm. To demonstrate the robustness and versatility of PoPS algorithm, we tested PoPS using four different environments from the OpenAI gym library\cite{openai}, each has a different model setting. The experiments were executed on ASUS Turbo GeForce GTX 1080 TI GPU. 
We compared the following algorithms: (i) the commonly used magnitude-based weight pruning \cite{pruning} using the gradual pruning framework \cite{pruneframework}, dubbed Magnitude-Base Gradual Pruning (MBGP), which is known to achieve strong performance in many DNN architectures; (ii) the Knowledge Distillation-Based Pruning (KDBP) algorithm, which demonstrated improvements in the pruning procedure in various classification settings by using knowledge distillation when pruning DNNs \cite{yu2017accelerating}; and (iii) the proposed PoPS algorithm. Note that the existing MBGP and KDBP vanilla versions conduct pruning in supervised learning settings. Thus, we adapted them to the DRL setting. As explained in Section \ref{ssec:popsarch}, two common DRL training methods were implemented, namely DQN, and Actor-Critic. Since the ground truth is not given in the DRL domain, we replaced it by the learned values. In the DQN setting, we used the TD values, given in (\ref{td_value}) as the ground truth during the fine-tuning procedure. In the Actor-Critic setting, we used the Advantage values, given in (\ref{eq:advantage_value}), to prune the actor, which is the DRL model used for the inference task (where the critic was trained using the TD values given in (\ref{td_value})). When implementing MBGP in the DRL domain, the pruned model interacts with the environment and feeds the experience replay with experiences. The fine-tuning procedure is preformed at each time-step by evaluating the TD or Advantage values using a batch of experiences from the experience replay.

When implementing KDBP in supervised learning settings, the student network is trained with the soft teacher output to improve the fine-tuning process, where the true labels are used to supervise the student output. The loss is computed using the cross-entropy between the true labels and the student output distribution, and the cross-entropy between the soft student output distribution and the soft teacher output distribution \cite{yu2017accelerating}. Therefore, when implementing KDBP in the DRL domain, we use the soft output teacher to improve the fine-tuning process. The pruned network interacts with the environment to obtain experiences which are then used to evaluate the distribution over TD values or Advantage values to compute the cross-entropy between those values and the student output distribution. They are used by the teacher as well to evaluate the soft teacher output distribution to compute the cross-entropy between the soft student output distribution and the soft teacher output distribution.

In both KDBP and MBGP, the interactions of the pruned agent with the environment makes the training less stable.

\subsection{Solving the Cartpole Environment with Compact Representations of a DQN}

\begin{figure}[htbp]
\centering \epsfig{file=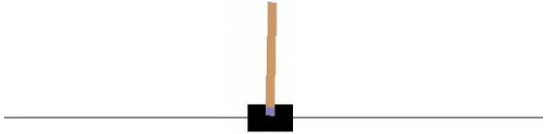,
width=0.5\textwidth}
\caption{An illustration of the Cartpole environment.}
\label{fig:avg_rate}
\end{figure}

In the Cartpole environment, a pole is attached to a cart, which is allowed to move along the x-axis, as illustrated in Fig. \ref{fig:avg_rate}. The
state space contains the velocity and position of the cart, the velocity of the tip of the pole, and the angle between the pole and the vertical line. 
At each time step, the agent is allowed to push the cart to the left or to the right, with a force of $-1$ and $+1$ unit, respectively, i.e., \(|\mathcal{A}|=2\). The pendulum starts upright, and the goal is to prevent it from falling over. At each time step where the pole remains upright, the agent receives a reward of $+1$. The episode ends when the pole is more the $15$ degrees from vertical, or the cart moves more then $2.4$ units from the center, or if the pole stays upright for $200$ time steps. The maximum possible reward
over an episode is thus $200$. A policy that solves the environment is considered to be a policy that achieves more than $195$ points in average over $100$ episodes.

The DRL architecture that was used in this experiment was a DQN with $3$ feed-forward fully connected layers with $256$, $256$, and $128$ neurons, respectively, summing up to a total of roughly $100K$ parameters. The initial model was trained using a target DQN, as described in \ref{ssec:popsarch}, until it was able to solve the environment. Then, we tested the performance of PoPS, MBGP, and KDBP algorithms in terms of solving the environment while obtaining compact representations of the DQN setting. 

We present the performance of PoPS algorithm in Table \ref{tab:pops_cartpole}, and MBGP and KDBP algorithms in Table \ref{tab:mbgp_cartpole}. It can be seen that PoPS algorithm achieves tremendous performance gain over the MBGP and KDBP algorithms after only several iterations. \emph{Specifically, PoPS generated a compact representation of the DRL model that solves the environment with a size of less than $0.3\%$ of the initial DQN representation size. By contrast, the MBGP and KDBP algorithms were not able to generate a representation of the DRL model that solves the environment with a size of less than $40\%$ of the initial DQN representation size.}

\begin{table*}[t]
\parbox{.45\linewidth}{
\caption{PoPS Performance in the Cartpole Environment}
\label{tab:pops_cartpole}
\centering
\small\begin{tabular}{|c|c|c|}
			\hline Iteration & \shortstack{$\#$ non-zero parameters  \\ (percentage of initial size)} & Average score \\  \hline
             0  & 100K (100\%) & $>$ 195  \\ \hline
             1  &  16K (16.2\%) & $>$ 195  \\ \hline
             2   & 2K (1.92\%) & $>$ 195   \\ \hline
             3  &  1.5K (1.5\%) & $>$ 195   \\ \hline
             4  &  1.2K (1.2\%) &  $>$ 195   \\ \hline
             5  &  908 (0.91\%) & $>$ 195  \\ \hline
             6  &  590 (0.59\%)  & $>$ 195   \\ \hline
             7  &  437 (0.43\%) & $>$ 195   \\ \hline
             8  &  292 (0.29\%) & $>$ 195  \\ \hline
             \textbf{9 (output)} &  \textbf{249 (0.25\%)} & \textbf{199.96} \\ \hline
    \end{tabular}
}
\hfill
\parbox{.45\linewidth}{
\caption{Performance of MBGP and KDBP in the Cartpole Environment}
\label{tab:mbgp_cartpole}
\centering
	\small\begin{tabular}{|c|c|c|}
			\hline \shortstack{$\#$ non-zero parameters  \\ (percentage of initial size)}  & \shortstack{Average score\\under MBGP} & \shortstack{Average score\\under KDBP}  \\  \hline
             100K (100\%) & $>$195 & $>$195 \\ \hline
             90K (90\%) & $>$195  & $>$195\\ \hline
             80K (80\%) & $>$195   & $>$195\\ \hline
             70K (70\%)  & $>$195  & $>$195 \\ \hline
             60K (60\%)  & $>$195 & $>$195  \\ \hline
             50K (50\%)  & $>$195 & $>$195 \\ \hline
             40K (40\%) & 180  & 130 \\ \hline
             35K (35\%)  & 140 & 164 \\ \hline
            30K (30\%) & $<$ \mbox{80} & 187\\ \hline
            $<$ \mbox{28K} (\mbox{28\%}) & $<$ \mbox{80} & $<$ \mbox{80}\\ \hline
    \end{tabular}
}
\end{table*}

\subsection{Solving the Lunar Lander Environment with Compact Representations of an Actor-Critic Architecture}

\begin{figure}[htbp]
\centering \epsfig{file=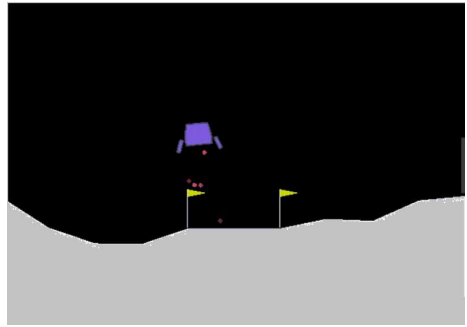,
width=0.4\textwidth}
\caption{An illustration of the Lunar lander environment.}
\label{fig:lunar_lander}
\end{figure}

In the second experiment, we tested the performance of the algorithms in the Lunar Lander environment. The environment consists of a ship and a landing pad in a shifting landscape, as illustrated in Fig. \ref{fig:lunar_lander}. The observation is an $8$ dimensional state vector that includes the ship's coordinates, velocity, lander-angle, angular-velocity, right-leg and left-leg grounded flag. The action space consists of $4$ discrete actions: do nothing, fire main engine, fire left engine, and fire right engine. Successful landing yields $100$ points, crash landing yields $-100$ points, each leg that contacts the ground yields $10$ points, and firing the main engine would subtract $0.3$ points per time step. An episode begins in a random state and ends when the lander lands successfully, crashes, flies out of bounds, or reaches $1000$ time steps. A policy that solves the environment is considered to be a policy that achieves more than $200$ points in average over $100$ episodes.

The architecture that was used in this experiment was an Actor-Critic network with $3$ feed-forward fully connected layers with $64$ neurons each, summing up to a total of roughly $9K$ parameters. The initial model was trained with a Critic in an Actor-Critic fashion until it was able to solve the environment. Then, we tested the performance of PoPS, MBGP, and KDBP algorithms in terms of solving the environment while obtaining compact representations of the Actor-Critic setting. 

We present the performance of PoPS algorithm in Table \ref{tab:pops_lunalander}, and MBGP and KDBP algorithms in Table \ref{tab:mbgp_lunalander}. It can be seen again that PoPS algorithm achieves tremendous performance gain over the MBGP and KDBP algorithms after only several iterations. \emph{Specifically, PoPS generated a compact representation of the DRL model that solves the environment with a size of less than $0.8\%$ of the initial Actor-Critic representation size. By contrast, the MBGP and KDBP algorithms were not able to generate a representation of the DRL model that solves the environment with a size of less than $68\%$ of the initial Actor-Critic representation size.} This experiment demonstrates the ability of PoPS to detect redundancy in a smaller model as well.

\begin{table*}[t]
\parbox{.45\linewidth}{
\caption{PoPS Performance in the Lunar Lander Environment}
\label{tab:pops_lunalander}
\centering
		\small\begin{tabular}{|c|c|c|}
			\hline Iteration & \shortstack{$\#$ non-zero parameters  \\ (percentage of initial size)} & Average score \\  \hline
             0 & 9K (100\%) & $>$ 200  \\ \hline
             1  & 1.4k (16.7\%)  &  $>$ 200  \\ \hline
             2  & 200 (2.22\%) & $>$ 200    \\ \hline
             3  & 160 (1.78\%) & $>$ 200   \\ \hline
             \textbf{4 (output)} &  \textbf{66 (0.73\%)} &  \textbf{249}   \\ \hline
    \end{tabular}
}
\hfill
\parbox{.45\linewidth}{
\caption{Performance of MBGP and KDBP in the Lunar Lander Environment}
\label{tab:mbgp_lunalander}
\centering
	\small\begin{tabular}{|c|c|c|}
			\hline \shortstack{$\#$ non-zero parameters  \\ (percentage of initial size)}  & \shortstack{Average score\\under MBGP} & \shortstack{Average score\\under KDBP}  \\  \hline
            9K (100\%)   & $>$ 200 & $>$ 200 \\ \hline
            8.1K (90\%)     & $>$ 200 & $>$ 200  \\ \hline
            7.2K (80\%) & $<$ \mbox{100} & $>$ 200  \\ \hline
            6.2K (70\%)     & $<$ 100 & $>$ 200  \\ \hline
            6.1K (68.6\%)     & $<$ 100 & 120  \\ \hline
            $<$ 6K (68\%)     & $<$ 100 & $<$ 100  \\ \hline         
    \end{tabular}
}
\end{table*}

\begin{table*}[t]
\parbox{.45\linewidth}{
\caption{PoPS Performance in the Pong Environment}
\label{tab:pops_pong}
\centering
		\small\begin{tabular}{|c|c|c|}
			\hline Iteration & \shortstack{$\#$ non-zero parameters  \\ (percentage of initial size)} & Average score \\  \hline
             0 & 1.6M (100\%) & $>$18  \\ \hline
             1       &  266K (15.8\%)  &  $>$18  \\ \hline
             2  &  128K (7.6\%) & $>$18   \\ \hline
             3  & 60K (3.6\%) & $>$18   \\ \hline
             4  &  22K (1.3\%) &  $>$18   \\ \hline
             \textbf{5 (output)}  &  \textbf{15K (0.92\%)} & \textbf{18.9}  \\ \hline
    \end{tabular}
}
\hfill
\parbox{.45\linewidth}{
\caption{Performance of MBGP and KDBP in the Pong Environment}
\label{tab:mbgp_pong}
\centering
	\small\begin{tabular}{|c|c|c|}
			\hline \shortstack{$\#$ non-zero parameters  \\ (percentage of initial size)}  & \shortstack{Average score\\under MBGP} & \shortstack{Average score\\under KDBP}  \\  \hline
             1.6M (100\%)    & $>$ 18 & $>$ 18 \\ \hline
             1.28M (80\%)   & $>$ 18  & $>$ 18 \\ \hline
             0.96M (60\%)   & $<$ 0 & $>$ 18   \\ \hline
             0.64M (40\%) & $<$ 0 & $>$ 18\\ \hline
             0.34M (20\%) & $<$ 0 & $>$ 18\\ \hline
             0.1M (6\%) & $<$ 0 & 16\\
             \hline
             $<$ 0.08M (5\%) & $<$ 0 & $<$ 0\\ \hline
    \end{tabular}
}
\end{table*}

\begin{table*}[t]
\parbox{.45\linewidth}{
\caption{PoPS Performance in the PACMAN Environment}
\label{tab:pops_pacman}
\centering
		\small\begin{tabular}{|c|c|c|}
			\hline Iteration & \shortstack{$\#$ non-zero parameters  \\ (percentage of initial size)} & Average score \\  \hline
             0 & 1.6M (100\%) & $> $1800  \\ \hline
             1 & 0.128M (8\%) & $> $1800  \\ \hline
             2 & 0.03M (1.9\%) & $> $1800  \\ \hline
             \textbf{3 (output)}  &  \textbf{4K (0.26\%)} & \textbf{2050}  \\ \hline
    \end{tabular}
}
\hfill
\parbox{.45\linewidth}{
\caption{Performance of MBGP and KDBP in the Pacman Environment}
\label{tab:mbgp_pacman}
\centering
	\small\begin{tabular}{|c|c|c|}
			\hline \shortstack{$\#$ non-zero parameters  \\ (percentage of initial size)}  & \shortstack{Average score\\under MBGP} & \shortstack{Average score\\under KDBP}  \\  \hline
             1.6M (100\%)    & $>$ 1800 &   $>$ 1800 \\ \hline
             1.28M (80\%)   & $>$ 1800  &   $>$ 1800 \\ \hline
             0.96M (60\%)   & $>$ 1800 &   $>$ 1800   \\ \hline
             0.64M (40\%) & $>$ 1800 &  $>$ 1800\\ \hline
             0.176M (11\%) & $>$ 1800  &   $>$ 1800\\ \hline
             0.174M (10.9\%) &  1020 &  $>$ 1800\\ \hline
             0.16M (10\%) & $<$ 500 &  $>$ 1800\\ \hline
             0.128M (8\%) & $<$ 500 &  1550\\ \hline
             0.12M (7.5\%) & $<$ 500 &  $<$ 500\\ \hline
    \end{tabular}
}
\end{table*}

\subsection{Solving the Pong Environment with Compact Representations of a Convolutional Neural Network (CNN)}

\begin{figure}[htbp]
\centering \epsfig{file=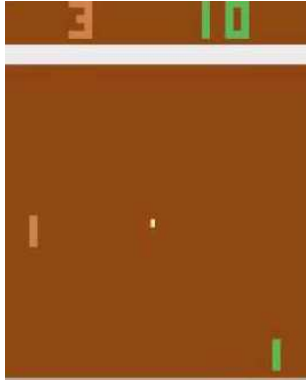,
width=0.35\textwidth}
\caption{An illustration of the Pong environment.}
\label{fig:pong}
\end{figure}

In the third experiment, we tested the performance of the algorithms in the Pong environment. Pong is a two-dimensional game that simulates a tennis table. The player uses the paddle to hit a ball back and forth, and it aims at passing the ball beyond the opponent's pad. An illustration is given in Fig. \ref{fig:pong}. The observation is an RGB image of the screen which is an array of shape $(210,160,3)$. The state space is comprised of $4$ grey scaled frames of size $84X84$ stacked together. The action space consists of $3$ discrete actions: up, down, and stay. A reward of $+1$ is received when the player passes the ball beyond the opponent's pad, whereas a reward of $-1$ is received when the ball passes the player's pad. The episode ends when a score of $21$ (winning ) or $-21$ (losing) is reached. A policy that solves the environment for the player is considered to be a policy that achieves more than $18$ points in average over $100$ episodes.

The Pong environment has a very high dimension. Thus, the capacity should not be too small, to avoid missing important experiences and suffering from a performance reduction. In PoPS, the teacher feeds batches of experiences before each fine-tuning or training phase by interacting with the Pong environment. Thus, the experiences in the experience replay are shifted dynamically, giving the pruned model the notion that there is a limitless amount of experiences. This method covers the large state space of Pong environment efficiently.

The architecture that was used in this experiment was a CNN with $3$ convolutional layers with a filter size of $8, 4$, and $3$, respectively, and the numbers of channels were $32, 64$, and $64$, respectively. The $3$ convolutional layers were followed by one fully connected layer with $512$ neurons. The initial model was trained using a target DQN, as described in Section \ref{ssec:popsarch}. 
The model was trained until it was able to solve the environment. Then, we tested the performance of PoPS, MBGP, and KDBP algorithms in terms of solving the environment while obtaining compact representations of the DQN with the CNN setting.

We present the performance of PoPS algorithm in Table \ref{tab:pops_pong}, and MBGP and KDBP algorithms in Table \ref{tab:mbgp_pong}. It can be seen again that PoPS algorithm achieves tremendous performance gain over the MBGP and KDBP algorithms after only several iterations. \emph{Specifically, PoPS generated a compact representation of the DRL model that solves the environment with a size of less than $1\%$ of the initial DQN with CNN representation size. By contrast, the KDBP algorithm, although presented a significant improvement as compared to the first two experiments, was not able to generate a representation of the DRL model that solves the environment with a size of less than $6\%$ of the initial DQN with CNN representation size. The MBGP algorithm performed poorly in this setting and was not able to generate a representation of the DRL model that solves the environment with a size of less than $60\%$ of the initial DQN with CNN representation size.} 

\subsection{Solving the Pacman Environment with Compact Representations of a Dueling DQN with CNN}
\begin{figure}[htbp]
\centering \epsfig{file=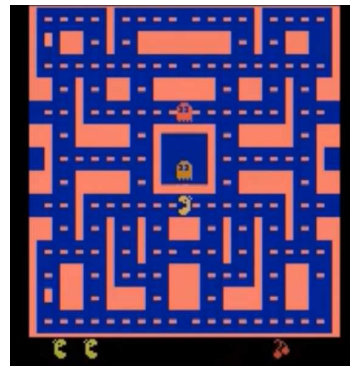,
width=0.35\textwidth}
\caption{An illustration of the Pacman environment.}
\label{fig:pacman}
\end{figure}

In the fourth experiment, we tested the performance of the algorithms in the Pacman environment. The player (i.e., agent) controls the Pacman and tries to collect all the Pac-Dots while avoiding the ghosts. An illustration is given in Fig. \ref{fig:pacman}. The observation is an RGB image of the screen which is an array of shape $(210,160,3)$. The state space is comprised of $4$ grey scaled frames of size $84X84$ stacked together. The action space consists of $9$ discrete actions: none, up, down, right, left, up-right, up-left, down-left and down-right. A reward of $+10$ is given for each collection of a Pac-Dot, $+50$ for Power Pellet and $+200$ for destroying vulnerable ghosts (multiplied by two if the player destroys them in succession). The episode ends once the Pacman runs out of lives or if the Pacman manages to collect all the Pac-Dots. A policy that solves the environment for the player is considered to be a policy that achieves more than $1800$ points in average over $100$ episodes.

The architecture that we used in this experiment is a CNN with $3$ convolutional layers with a filter size of $8, 4$, and $3$, respectively, and the numbers of channels were $32, 64$, and $64$, respectively. The $3$ convolutional layers were followed by one fully connected layer with $512$ neurons and then an output layer with $10$ neurons to implement a dueling architecture \cite{wang2015dueling}. The initial model was trained using a target dueling DQN, as described in Section \ref{ssec:popsarch}. 
The model was trained until it was able to solve the environment. Then, we tested the performance of PoPS, MBGP, and KDBP algorithms in terms of solving the environment while obtaining compact representations of the dueling DQN with the CNN setting.

We present the performance of PoPS algorithm in Table \ref{tab:pops_pacman}, and MBGP and KDBP algorithms in Table \ref{tab:mbgp_pacman}. It can be seen that PoPS achieves tremendous performance gain over the MBGP and KDBP algorithms in this setting as well after only several iterations. \emph{Specifically, PoPS generated a compact representation of the DRL model that solves the environment with a size of less than $0.3\%$ of the initial dueling DQN with CNN representation size. By contrast, the MBGP and KDBP algorithms were not able to generate a representation of the DRL model that solves the environment with a size of less than $10\%$, and $8\%$, respectively, of the initial dueling DQN with CNN representation size.}

\section{Conclusion}
\label{sec:conclusion}

We have developed a policy pruning and shrinking (PoPS) framework for achieving compact representations of DNNs in the DRL domain. The suggested framework is based on a novel iterative policy pruning and shrinking method that leverages the power of transfer learning when training the DRL model. Extensive experimental results using Cartpole, Lunar Lander, Pong, and Pacman environments demonstrated tremendous performance gain of PoPS over the MBGP and KDBP algorithms in all four environments and different model settings. The results obtained by PoPS present the state-of-the-art performance in terms of training DRL models with strong required performance while minimizing the size of the DNN in the DRL domain. This research demonstrates the great potential of PoPS in making DRL models practically appealing for a wide range of applications that use systems with limited hardware resources. We developed open source software of PoPS framework which is available in \cite{githubpops}. Researchers and developers in related fields are invited to incorporate PoPS in their working environment. 
Finally, despite the success of DRL-based algorithms in many fields, not much is known about their theoretical performance guarantees. In particular, a potential future direction is to establish theoretical guarantees for compression ratios in the DRL domain.

\section{Appendix}

\subsection{Implementation Details of the Open Source Software}

\begin{figure}[htbp]
\centering \epsfig{file=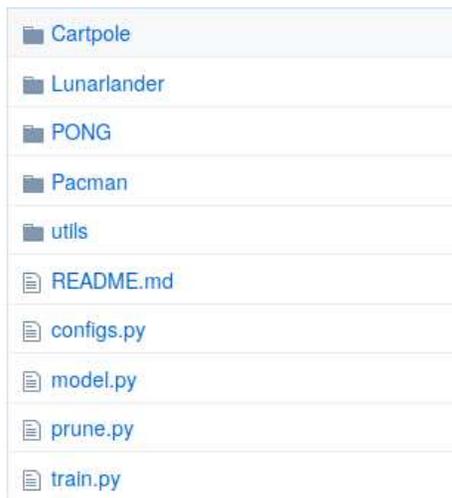,
width=0.5\textwidth}
\caption{The PoPS project directory as seen in GitHub.}
\label{fig:project}
\end{figure}

In this appendix, we summarize the implementation details of the open source software that we developed. PoPS was developed using Python and is available at GitHub (see link in \cite{githubpops}). Fig. \ref{fig:project} illustrates the top level hierarchy of all the modules in the PoPS project. Interested readers are encouraged to read the README file for further information.

1) The \textit{configs.py} file holds the configuration for each environment and model, parameters (such as the target sparsity and the pruning frequency for the pruning procedure) that affect the initial training phase, and the PoPS procedure.

2) The \textit{model.py} file contains the model architecture for each environment such as class DQNPong, class CartPoleDQN, class ActorLunarLander, and CriticLunarLander. These models are used for the initial training phase, and follow the DQNAgent interface. Each model is associated with a Student version that inherits it, such as StudentPong. The Student version is adjusted for the PoPS algorithm such that the loss function is the loss presented in (\ref{kldiv})  and the architecture is a dynamic architecture which is defined by the redundancy measures.

3) The \textit{train.py} file contains functions that execute the policy distillation training procedure as well as the IPP's pruning and fine-tuning steps. The functions are well documented and are used by a variety of models and environments.

4) The file \textit{prune.py} contains the IPP module orchestrating the pruning phase in the PoPS algorithm as detailed in Section \ref{sec:PoPS} with the \emph{train\_student} function which is used for pruning and fine tuning the model.

5) The \textit{utils} package contains several modules with helpful utilities. 

\begin{figure}[htbp]
\centering \epsfig{file=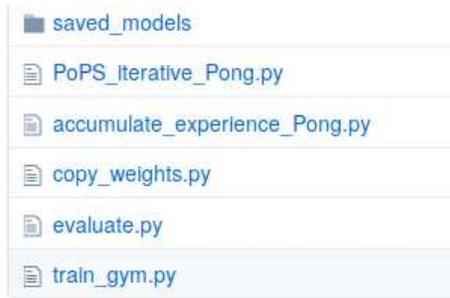,
width=0.5\textwidth}
\caption{The experiment directory of Pong as seen in GitHub.}
\label{fig:pong_git}
\end{figure}

For each environment there is a unique evaluation tool and an accumulated experience function. These two functions with the modules described above implement the interface needed for the PoPS algorithm. The implementation of PoPS algorithm is similar for each environment and is well documented at GitHub, including the experiments presented in Section \ref{sec:experiemnts}. For each environment, the initial trained model is given in the \emph{saved\_models} folder as well, with a train script for the initial training phase for reconstructing the experiments from scratch if desired. For instance, a description of the Pong experiment is given below. An illustration of the experiment directory as seen in GitHub is given in Fig. \ref{fig:pong_git}.

\begin{itemize}
\item The \textit{evaluate.py} file is the unique evaluation tool for the Pong environment. It can used as a stand alone script as well to evaluate models in the Pong environment.
\item The \textit{copy\_weights.py} file is the function needed for PoPS to create the initial \(Q_0\) Student model from the teacher model \(Q^{teacher}\). It is a compatibility function for the purpose of creating a Student version of a model.
\item The \textit{train\_gym.py} file is a script to initially train the model to solve the environment.
\item The \textit{PoPS\_iterative\_Pong.py} file is for the PoPS algorithm which uses the modules discussed above. It is very easy to read and implement. The first step is to import the teacher model and then create the initial student model with \textit{copy\_weights.py}. Then, there is a convergence loop where PoPS prunes the model using IPP, evaluates the redundancy, creates a model with a size according to the measured redundancy, trains it, and repeats until convergence.
\item The \textit{accumulate\_experience\_Pong.py} file is a module which fills the Experience Replay with interactions of the teacher model with the Pong environment. The experiences in the Experience Replay are used later for training with respect to the KL divergence loss function (\ref{kldiv}).
\end{itemize}

\bibliographystyle{ieeetr} 

\end{document}